\documentclass[letterpaper, 10 pt, conference]{ieeeconf} 

\IEEEoverridecommandlockouts % This command is only needed if you want to use the \thanks command

\usepackage[english]{babel}
\usepackage[utf8x]{inputenc}
\usepackage{amsmath}
\DeclareMathOperator*{\argmin}{arg\,min}
\usepackage{graphicx}
\usepackage[colorinlistoftodos]{todonotes}
\usepackage{cite}
\usepackage{algorithm}
\usepackage{algorithmic}
\usepackage{hyperref}
\usepackage{amsfonts}
\usepackage{booktabs, float}
\title{A Point Cloud Completion Approach for the Grasping of Partially Occluded Objects and Its Applications in Robotic Strawberry Harvesting}
\author{
Ali Abouzeid$^{1}$, 
    Malak Mansour$^{1}$, 
    Chengsong Hu$^{1}$, 
    Dezhen Song$^{1}$ % <-this % stops a space  
\thanks{$^{1}$Authors are with the Department of Robotics,
        Mohamed bin Zayed University of Artificial Intelligence, Masdar City, Abu Dhabi, UAE. Emails:\tt\small \{ali.abouzeid, malak.mansour, chengsong.hu, dezhen.song\}@mbzuai.ac.ae
}
}

\begin{document}
\maketitle

\begin{abstract}
In robotic fruit picking applications, managing object occlusion in unstructured settings poses a substantial challenge for designing grasping algorithms. Using strawberry harvesting as a case study, we present an end-to-end framework for effective object detection, segmentation, and grasp planning to tackle this issue caused by partially occluded objects. Our strategy begins with point cloud denoising and segmentation to accurately locate fruits. To compensate for incomplete scans due to occlusion, we apply a point cloud completion model to create a dense 3D reconstruction of the strawberries. The target selection focuses on ripe strawberries while categorizing others as obstacles, followed by converting the refined point cloud into an occupancy map for collision-aware motion planning. 
Our experimental results demonstrate high shape reconstruction accuracy, with the lowest Chamfer Distance compared to state-of-the-art methods with 1.10 mm, and significantly improved grasp success rates of 79.17\%, yielding an overall success-to-attempt ratio of 89.58\% in real-world strawberry harvesting. Additionally, our method reduces the obstacle hit rate from 43.33\% to 13.95\%, highlighting its effectiveness in improving both grasp quality and safety compared to prior approaches.
%Our experimental findings show high segmentation accuracy and enhanced grasp success rates in simulations and actual environments, surpassing current models, and confirming the efficacy of our approach. 
This pipeline substantially improves autonomous strawberry harvesting, advancing more efficient and reliable robotic fruit picking systems.\\
Code is available at \url{https://github.com/Malak-Mansour/PointCloud_Completion_for_Grasping}
\end{abstract}

\section{Introduction}

\begin{figure}[t]
    \centering    \includegraphics[width=\linewidth]{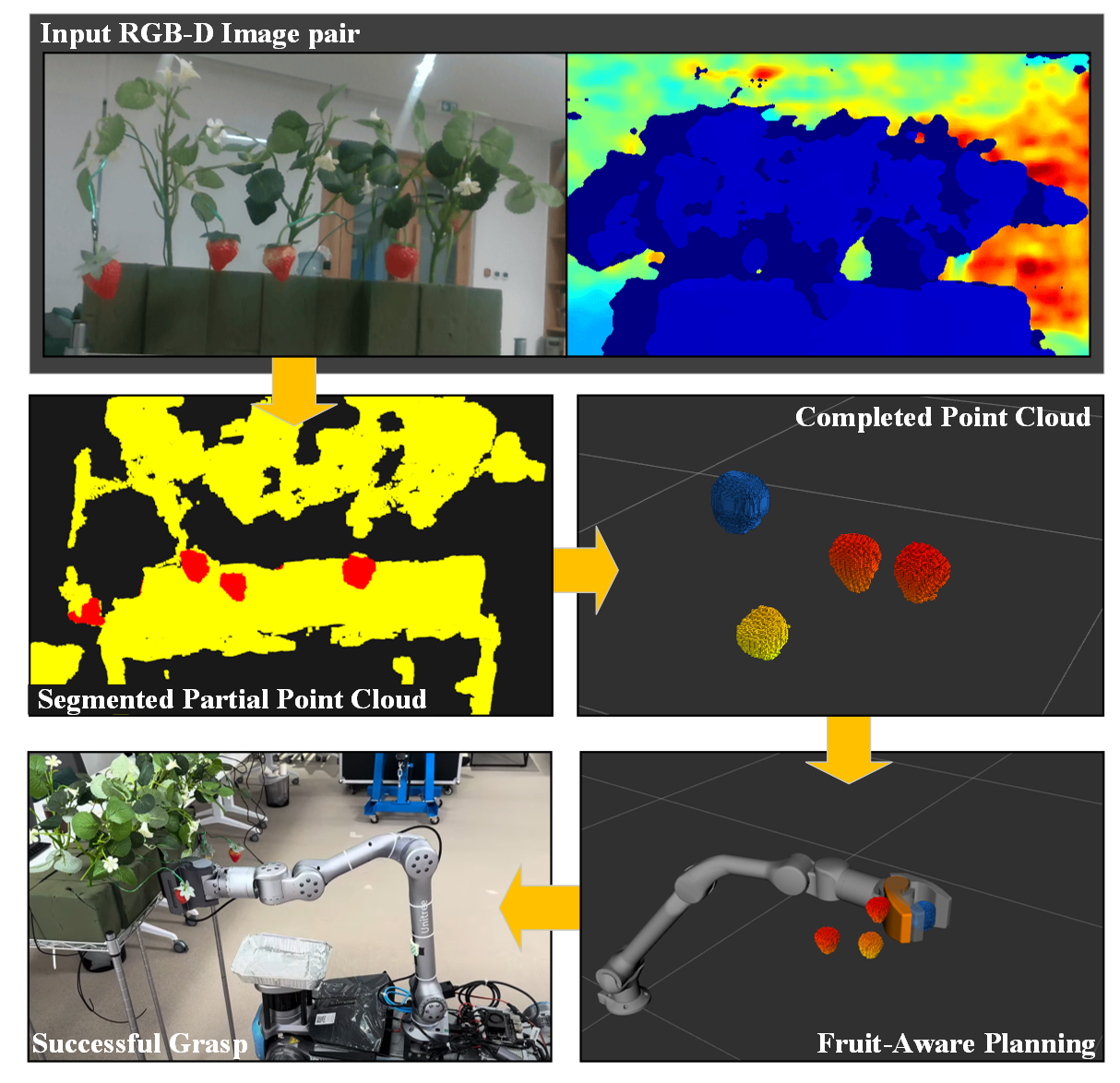}
    \caption{Top: Input RGB-D image pair showing a multiple strawberry scene. Second row: (left) Segmented point cloud of strawberries, with red points indicating the segmented regions; (right) completed strawberry point cloud after processing the segmented data through a completion model. Bottom row: (right) Planning scene with fruit-aware path planning, where blue points represent the target strawberry and orange/yellow points are treated as obstacles; (left) successful grasping of the target strawberry by the robotic arm using the proposed pipeline}
    \label{fig:robot_with_plant}
\end{figure}

Strawberries are a globally popular fruit with significant commercial value, but their cultivation is labor-intensive, particularly during harvesting, which accounts for a substantial portion of production costs \cite{guo2024technology}. The strawberry industry faces ongoing challenges, including seasonal labor shortages and rising expenses, which have increased interest in automation as a potential solution \cite{ryan2023labour}. The compact size of strawberry plants and the visually distinct appearance of their fruit make them well-suited for robotic harvesting, an area that has attracted growing research interest.

Despite this potential, automating strawberry harvesting presents several challenges, including the fragility of the fruit, which requires gentle, damage-free handling to maintain quality, and the accurate perception of berries in complex, unstructured field environments \cite{wang2024review}. Compounding these perception difficulties, robotic systems often struggle to handle partially occluded objects, an issue that remains relatively under-explored in the field \cite{zhou2022intelligent}. Unlike industrial pick-and-place tasks, where objects are fully visible and structured environments aid perception, strawberry harvesting occurs in cluttered settings, where occlusions are common due to leaves, stems, and other fruit, further complicating robotic perception and manipulation.

To enable reliable whole-fruit grasping, a robotic system must acquire detailed semantic and geometric information about target strawberries, even under partial visibility. Here, point cloud completion plays a pivotal role: By reconstructing occluded regions of the fruit, it provides a comprehensive 3D representation of the strawberry, enabling robust grasp planning. This approach not only ensures precise alignment of the gripper to avoid damaging the target fruit through misalignment or excessive force but also identifies and localizes non-target strawberries (e.g., unripe or occluded fruits) that must be avoided during manipulation. By completing partial point clouds of the scene, the system reconstructs both the target’s geometry and the surrounding obstacles, bridging the gap between partial sensory input and the complete spatial understanding required for safe, effective autonomous harvesting in real-world conditions. However, a significant hurdle remains: the sim-to-real gap, particularly in handling noisy RGB-D camera data \cite{ma2024sim, weibel2019addressing}. Existing methods often fail to directly address these discrepancies, limiting their real-world applicability. Fig. \ref{fig:robot_with_plant} demonstrates how our robotic arm successfully grasps the target strawberry without damaging the surrounding strawberries, due to the accurate completion of the partial strawberries' point clouds.

% The main contribution of this paper is a novel approach for robotic strawberry harvesting that combines three key steps to achieve reliable, collision-free operation in cluttered environments. Our method addresses these challenges by: (1) denoising and segmenting raw point clouds to improve input data quality, directly tackling the sim-to-real gap; (2) reconstructing complete 3D strawberry models using point cloud completion to overcome occlusions and limited viewpoints; and (3) selecting targets and planning collision-free paths to ensure the robot grasps only ripe strawberries while avoiding damage to the plant and nearby fruits. By integrating these steps, our system improves robustness to sensor noise and environmental clutter, enabling safer and more efficient harvesting in real-world settings. 

Our main contributions are summarized as follows:
\begin{itemize}
    \item Robust point cloud processing that reduces sensor noise and bridges the sim-to-real gap through advanced denoising and segmentation.
    \item Complete 3D reconstruction of occluded or partially visible strawberries, overcoming the limitations of single-view camera captures.
    \item Accurate grasping of ripe strawberries by treating non-target strawberries as obstacles to avoid damaging them.
\end{itemize}
Our novel robotic strawberry harvesting pipeline significantly outperforms existing methods in cluttered, real-world environments by addressing these key challenges. Extensive experiments demonstrate significant improvements in grasp success rates and point cloud completion accuracy compared to state-of-the-art methods, as well as a notable reduction in hit rates with non-target strawberries, highlighting the effectiveness of our approach.

% The proposed pipeline enhances robustness to noise and occlusions. In sum, we make three key claims: our approach is able to (i) directly tackle the sim-to-real challenges; (ii) mitigate occlusion and limited viewpoint issues; (iii) achieve robust, collision-free, and damage-free manipulation in cluttered environments.

% \subsection{Contributions:}
% \begin{itemize}
%     \item Pipeline for pose estimation that maximizes pose accuracy while considering the real-life environment with obstacles and unclear strawberry views
%     \item Evaluating the sim-to-real gap using simulation and real-life datasets on the same pipeline
%     \item Segmenting point clouds using image segmentation masks
%     \item Completing the point cloud of strawberries before estimating the pose for more robustness and completion
%     \item Treating leaves and none-target strawberries as obstacles during trajectory planning
%     % \item  Accuracy metric of the Point cloud completion model 
% \end{itemize}

\begin{figure*}[h!]
    \centering
    \includegraphics[width=\textwidth]{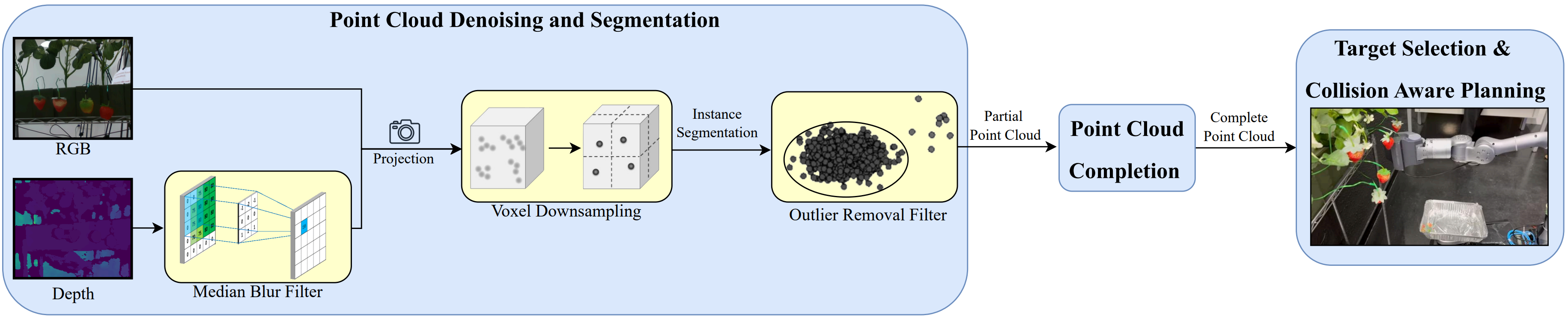}
    
    \caption{Our robotic strawberry harvesting pipeline consists of: Point Cloud Denoising and Segmentation, Point Cloud Completion \cite{PointAttn}, and Target Selection with Collision-Aware Planning, where non-target strawberries are treated as obstacles.}
    \label{fig:pipeline}
\end{figure*}

\section{Related Work}

\subsection{Agricultural and Harvesting Robots} 

Robotic strawberry harvesting has been widely explored to improve labor efficiency and precision. It has been approached primarily through two methods: peduncle cutting \cite{parsa2022peduncle, 3d_shape_completion, parsa2024modular} and whole-fruit grasping \cite{uppalapati2020berry}. While peduncle cutting is a widely adopted method for robotic strawberry harvesting, whole-fruit grasping offers distinct advantages in cluttered environments where occlusions and structural complexity are prevalent. In such settings, leaves, neighboring fruits, and plant structures often obscure the peduncle, demanding highly precise localization and tool maneuverability to execute cuts— a task complicated by limited visibility and spatial constraints \cite{zhang2020state}. This challenge highlights the need for robust perception and grasp planning strategies. Whole-fruit grasping, in contrast, directly addresses these limitations by leveraging the fruit’s larger and more accessible surface area, reducing reliance on pinpoint accuracy.

A key challenge in robotic harvesting is ensuring the visibility of the fruits for accurate detection and manipulation, addressed by leaf manipulation techniques, such as those proposed in \cite{leaf_manipulation, gursoy2024occlusion}, which displaces occluding leaves to enhance fruit visibility and improve shape and pose estimation for robotic systems. Furthermore, precise grasp point detection and fruit orientation estimation are essential for effective harvesting, achieved through methods like 3D visual detection that fuse color and geometry to identify graspable regions, such as sweet pepper peduncles for accurate robotic cutting \cite{grasp_point_detection}. Grasp point detection can be combined with pose estimation, leveraging either deep learning-based general feature extraction methods \cite{rotation_estimation, 6DOF_harvesting_IROS} or key points-based estimation methods to recover the main axis of fruits, as demonstrated in \cite{le2023key, zheng2021mango, tafuro2022strawberry}.

\subsection{3D Shape Completion}

Shape completion plays a crucial role in reconstructing occluded regions of objects, which is particularly important for applications like robotic harvesting. Various deep learning-based approaches have been developed to address general shape completion tasks. Methods such as DeepSDF \cite{park2019deepsdf} and \cite{cai2022learning} learn continuous shape representations that generalize across both known and unknown shapes, enabling robust shape inference. Extensions like MV-DeepSDF \cite{liu2023mv} further enhance this by leveraging multi-sweep point clouds for implicit modeling. Attention-based point cloud completion models like PointAttn \cite{PointAttn} have been designed to handle general object sets, though they often lack domain-specific adaptations. 

In the context of robotic harvesting, shape completion is essential for reconstructing occluded fruit regions to enhance grasping success. While general deep learning methods provide a foundation, they often fail to address the real-time requirements critical for robotic systems. Recent research has addressed these challenges through specialized architectures, particularly the work in \cite{magistri2022contrastive}, which employed transformer networks with contrastive learning and LiDAR priors to enable efficient, real-time completion of occluded fruit shapes.

Building upon these specialized shape completion approaches, a robotic fruit harvesting system was developed based on 3D shape completion with 6-DoF pose estimation for grasping in vertical farms \cite{3d_shape_completion}. This system outperforms traditional methods like grasp point detection, orientation estimation, and ellipsoid fitting \cite{grasp_point_detection, rotation_estimation} with higher success rates for strawberries and tomatoes. However, this approach overlooks obstacles around the target fruit, limiting performance in dense clusters. Ellipsoid fitting methods struggle with occlusions due to limited sample points \cite{ellipsoid_fitting}, while scene categorization techniques discard occluded fruits entirely \cite{discard_obstacles}. In contrast, our method enables robust fruit harvesting even in cluttered environments by addressing both occlusion and obstacle avoidance.

% \subsection{Obstacle Avoidance in Cluttered Environments} 

% Robotic harvesting in cluttered scenes requires precise distinction between target fruits and obstacles. Traditional methods, such as template-based detection (e.g., LINEMOD \cite{texture_less_3D_cluttered_objects}), improve object localization in complex environments. Prior works like \cite{3d_shape_completion} refine point cloud data to enhance obstacle representation, yet fail to account for real-time collision-aware grasp planning under occlusions, which we achieve. 

\subsection{Sim-to-Real Challenges in Robotic Perception}

Previous research has tackled the sim-to-real gap in robotic perception using various techniques. R2SGrasp \cite{cai2024real} employs a Real-to-Sim Data Repairer (R2SRepairer) to mitigate noise in real depth maps at the data level, enabling real-to-sim adaptation of camera noise. \cite{jang2024bridging} focuses on bridging the simulation-to-real gap for depth images in deep reinforcement learning applications. Other approaches leverage deep learning techniques, such as refinement and self-supervised domain adaptation, to address noise and domain discrepancies \cite{denoising}. While these deep learning methods are effective, they remain computationally expensive for real-time applications. In contrast, our work tackles the sim-to-real challenge in strawberry harvesting through lightweight filtering techniques, providing an efficient solution to sensor noise challenges in agricultural robotics.

%Our approach builds upon these efforts by explicitly completing occluded point clouds, estimating fruit centroids, and generating grasp poses that account for both shape and orientation variations.

\section{Problem Formulation}
Given a partially occluded view of the strawberry plant, our method completes the point cloud of the strawberries for accurate grasp planning while treating non-target strawberries as obstacles. Our approach is under the assumption that the strawberry is characterized by a prior fixed shape and size (CAD model) for both training and evaluation.

To formalize our approach, we define key notations and outline the system's input and output representations.

\textbf{\textit{Common notations:}}
\begin{description}
\item[$H$] Image height 
\item[$W$] Image width
\item[$I_{RGB}$] RGB image $\in \mathbb{R}^{H \times W \times 3}$ 
\item[$I_{D}$] Depth image $\in \mathbb{R}^{H \times W}$
\item[$\mathbf{p}$] Input point cloud
%\item[$F_{\theta}$] learned point cloud completion function
%\item[$\mathcal{M}$] function that converts the completed point clouds into a 3D occupancy grid
\end{description}

\textbf{\textit{Input:}} The system takes as input an RGB-D image pair, 
$I_{RGB}$ and $I_{D}$, capturing a front-facing view of the strawberry plant.

\textbf{\textit{Output:}} The system generates a completed 3D point cloud $\mathbf{p}_{i,\text{complete}} $ of the $i$-th strawberry from the partial denoised point cloud $\mathbf{p}_{i,\text{denoised}}$. These completed point clouds are then used for obstacle-aware motion planning, where the occupied space is represented as an occupancy map $M_{\text{occ}}$.

%\begin{equation}
    %\mathbf{p}_{i,\text{complete}} = F_{\theta}(\mathbf{p}_{i,\text{denoised}})
%\end{equation}

%\begin{equation}
    %M_{\text{occ}} = \mathcal{M} \left(\bigcup_{i} \mathbf{p}_{i,\text{complete}} \right)
%\end{equation}

\section{Algorithm}
Our pipeline processes RGB-D images of strawberry plants to enable accurate grasp planning. The approach consists of three main stages: (1) filtering and segmentation to isolate individual fruits, (2) point cloud completion to reconstruct the full 3D shape of each strawberry, and (3) target identification for grasp planning while treating other fruits as obstacles. Fig. \ref{fig:pipeline} illustrates the complete pipeline.

\subsection{Segmenting and Denoising Point Clouds}
To generate complete 3D models of individual strawberries from the input RGB-D pair ($I_{RGB}$ and $I_{D}$), we implement a multi-stage filtering process that ensures high-quality reconstruction.

We begin with depth image filtering to reduce sensor noise while preserving important edge features. The filtering operation $\mathcal{F}$ applies a median blur:
\begin{equation}
\begin{aligned}
    &I_{D,\text{filtered}}(u, v) = \\
    &\text{median} \left\{ I_D(u', v') \mid (u', v') \in \mathcal{N}_5(u, v) \right\},
\end{aligned}
\end{equation}
where \(\mathcal{N}_5(u, v)\) is the \(5 \times 5\) neighborhood centered at pixel \((u, v)\). The filtered depth image is:
\begin{equation}
    I_{D,\text{filtered}} = \mathcal{F}(I_D).
\end{equation}

Using \(I_{D,\text{filtered}}\), we generate a point cloud \(\mathbf{p}\) by projecting depth values into 3D space, incorporating RGB information from \(I_{\text{RGB}}\) and camera intrinsics \(K\):
\begin{equation}
    \mathbf{p} = \mathcal{E}(I_{\text{RGB}}, I_{D,\text{filtered}}, K).
\end{equation}

To reduce noise further, we apply voxel downsampling \(\mathcal{V}\) to \(\mathbf{p} \). Points are grouped into voxels of size \(v_s = 0.05\), and their positions are averaged within each voxel. Additionally, we enforce a minimum point threshold per voxel to filter noise:
\begin{equation}
    V_i = \left\{ p_j \in \mathbf{p} \mid \left\lfloor \frac{p_j}{v_s} \right\rfloor = i \right\},
\end{equation}
where \(i\) is the voxel index. The voxel center \(c_i\) is computed as:
\begin{equation}
    c_i = \frac{1}{|V_i|} \sum_{p_j \in V_i} p_j, \quad \text{if} \quad |V_i| \geq 30,
\end{equation}
otherwise \(c_i = (0, 0, 0)\). The downsampled point cloud \(\mathbf{p}_{\text{voxel}}\) is:
\begin{equation}
    \mathbf{p}_{\text{voxel}} = \mathcal{V}(\mathbf{p}) = \{ c_i \mid i \in \text{voxel indices} \}.
\end{equation}

Next, we perform instance segmentation on \(I_{\text{RGB}}\) to generate binary masks \(M = \{m_1, m_2, \dots, m_N\}\), where \(m_i \in \{0, 1\}^{H \times W}\) corresponds to an individual strawberry, and \(N\) is the number of detections. These masks are used to extract points from \(\mathbf{p}_{\text{voxel}}\):
\begin{equation}
    \mathbf{p}_i = \mathcal{X}(\mathbf{p}_{\text{voxel}}, m_i),
\end{equation}
where \(\mathcal{X}\) isolates points corresponding to mask \(m_i\).

Finally, to enhance the quality of each segmented point cloud, we apply an outlier removal filter \(\mathcal{O}\) to eliminate residual noise:
\begin{equation}
    \mathbf{p}_{i,\text{denoised}} = \mathcal{O}(\mathbf{p}_i).
\end{equation}

This stage reduces noise, isolates individual strawberries, and ensures high-quality point clouds for further processing.

% The outlier removal process generates a set of denoised point clouds $P_{i,denoised} \in \mathbb{R}^{n_{i,d} \times 3}$, where $n_{i,d}$ represents the number of points in the denoised point cloud for the $i$-th strawberry, with $i \in \{1,...,N\}$.

% By combining advanced depth filtering, voxel-based downsampling, segmentation, and outlier removal, this pipeline effectively generates clean and structured 3D models of strawberries for further processing in agricultural robotics applications.

\subsection{Point Cloud Completion}
In line with established approaches for shape completion, we employ a deep learning model based on \cite{PointAttn} to learn comprehensive shape representations of strawberries. The model processes a set of denoised partial point clouds and generates their complete counterparts through an encoder-decoder architecture, leveraging pre-trained weights obtained through training on a mix of simulated and real datasets.

The completion network $F_\theta$ with parameters $\theta$ transforms each denoised point cloud into its complete representation:

\begin{equation}
    \mathbf{p}_{i,complete} = F_\theta(\mathbf{p}_{i,denoised})
\end{equation}

During training, the model weights $\theta$ were optimized using a hierarchical loss function based on Chamfer Distance (CD). The network generates point clouds at different resolutions $(\mathbf{p}_0, \mathbf{p}_1, \mathbf{p}_2)$, where $\mathbf{p}_0$ represents a sparse seed point cloud, and $\mathbf{p}_1, \mathbf{p}_2$ are progressively denser completions. The Chamfer Distance between two point clouds $\mathbf{p}$ and $Q$ is defined as:

\begin{equation}
    d_{CD}(\mathbf{p},Q) = \sum_{p \in \mathbf{p}} \min_{q \in Q} ||p - q||_2^2 + \sum_{q \in Q} \min_{p \in \mathbf{p}} ||p - q||_2^2
\end{equation}

The total loss is computed against corresponding ground truth point clouds of matching densities:

\begin{equation}
    \mathcal{L} = \sum_{i=0}^2 \lambda_i d_{CD}(\mathbf{p}_i, S_i)
\end{equation}

where $\lambda_i$ are weighting coefficients and $S_i$ are the ground truth point clouds.

At inference time, we leverage the pre-trained weights for completing the denoised point cloud set. This process generates a set of complete point clouds $\mathbf{p}_{i,complete} \in \mathbb{R}^{n \times 3}$, where $n$ represents the number of points in each completed strawberry point cloud.

\subsection{Target Selection and Motion Planning}

Given the set of completed point clouds and their corresponding ripeness labels from the detection stage, we first filter for ripe strawberries to obtain the set of candidate targets $\{\mathbf{p}_{k,complete}\}_{k \in R}$, where $R$ is the index set of ripe strawberries. The optimal target is selected based on minimum Euclidean distance $k^*$ to the end-effector position $p_{ee}$:

\begin{equation}
    k^* = \argmin_{k \in R} ||c_k - p_{ee}||_2
\end{equation}

where $c_k$ is the centroid of the $k$-th completed point cloud.

The remaining point clouds, both ripe and unripe, are treated as obstacles:

\begin{equation}
    O = \bigcup_{i \neq k^*} \mathbf{p}_{i,complete}
\end{equation}

These obstacle point clouds are integrated into an occupancy map $M_{occ}$ for collision-aware inverse kinematics:

\begin{equation}
    M_{occ} = \mathcal{M}(O)
\end{equation}

where $\mathcal{M}$ converts the point clouds into a 3D occupancy grid. This allows for motion planning that avoids collisions with other strawberries while approaching the selected target.

Our pipeline is summarized more formally in Algorithm \ref{alg:strawberry_grasping}.

\begin{algorithm}[ht]
\caption{Strawberry Grasping Algorithm}
\label{alg:strawberry_grasping}
\begin{algorithmic}[1]
\STATE \textbf{Input:} $(I_{\text{RGB}}, I_D)$, $p_{ee}$
\STATE $\{m_i\}_{i=1}^n \gets \text{DetectRipeNonRipe}(I_{\text{RGB}})$
\STATE $\mathbf{p} \gets \text{GeneratePointCloud}(I_{\text{RGB}}, I_D, K)$
\STATE $\{\mathbf{p}_{i,\text{denoised}}\}_{i=1}^n \gets \text{DenoiseAndExtract}(\mathbf{p}, \{m_i\}_{i=1}^n)$
\STATE $\{\mathbf{p}_k\}_{k \in R} \gets \text{FilterRipe}(\{\mathbf{p}_{i,\text{denoised}}\}_{i=1}^n)$
\STATE $k^* \gets \text{SelectTarget}(\{\mathbf{p}_k\}_{k \in R}, p_{ee})$
\STATE $\{\mathbf{p}_o\} \gets \{\mathbf{p}_{i,\text{denoised}} \mid i \neq k^*\}$

\STATE $O \gets \emptyset$
\FOR{each $\mathbf{p}_{i,\text{denoised}}$}
    \STATE $\mathbf{p}_{i,\text{complete}} \gets \text{CompletePointCloud}(\mathbf{p}_{i,\text{denoised}})$
    \IF{$\mathbf{p}_{i,\text{denoised}} \in \{\mathbf{p}_o\}$}
        \STATE $O \gets O \cup \mathbf{p}_{i,\text{complete}}$
    \ENDIF
\ENDFOR

\STATE $M_{occ} \gets \text{GenerateOccupancyMap}(O)$
\STATE $G \gets \text{EstimateGrasp}(\mathbf{p}_{k^*,\text{complete}})$
\STATE $T \gets \text{PlanTrajectory}(G, M_{occ})$
\IF{$T$ is feasible}
    \STATE $\text{ExecuteGrasp}(T)$
    \STATE \textbf{return} Success
\ENDIF

\STATE \textbf{return} Failure
\end{algorithmic}
\end{algorithm}

\section{Experimental Setup}

\subsection{Data Collection}
We used a mixture of simulation and real-world data to train our completion model. For simulation data, we utilized NVIDIA Isaac Sim to create a virtual strawberry field with realistic occlusions. We simulated an Intel RealSense D435i RGB-D camera from the robot's perspective to capture partial point clouds of strawberries. This setup allowed us to simultaneously obtain ground truth completed point clouds for training and evaluation.

\begin{figure*}[!htbp]
    \centering
    \begin{minipage}[b]{0.48\textwidth}
        \centering
        \includegraphics[width=\textwidth,height=0.75\textwidth]{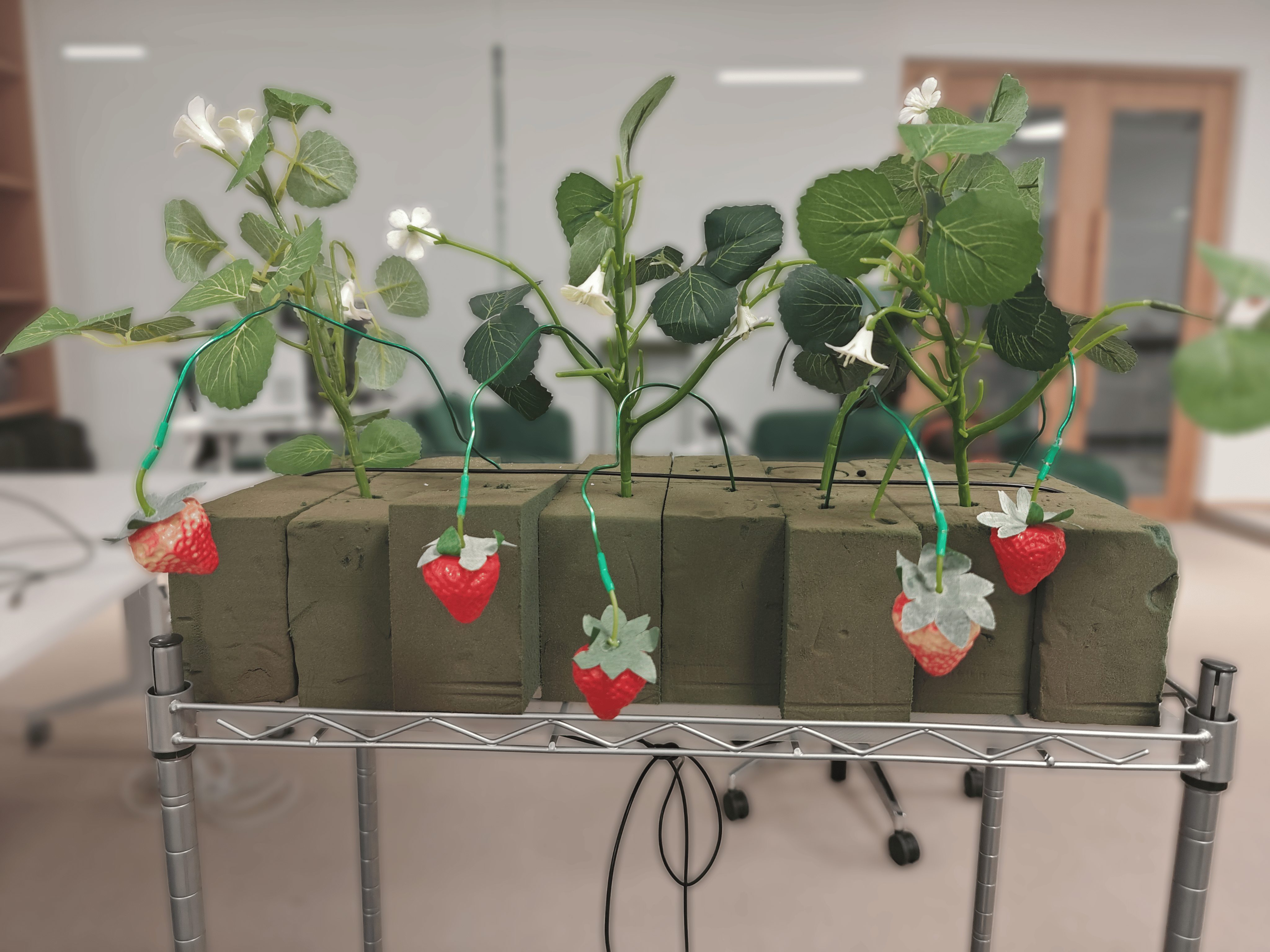} % Replace with actual real-world image file
    \end{minipage}
    \hfill
    \begin{minipage}[b]{0.48\textwidth}
        \centering
        \includegraphics[width=\textwidth,height=0.75\textwidth]{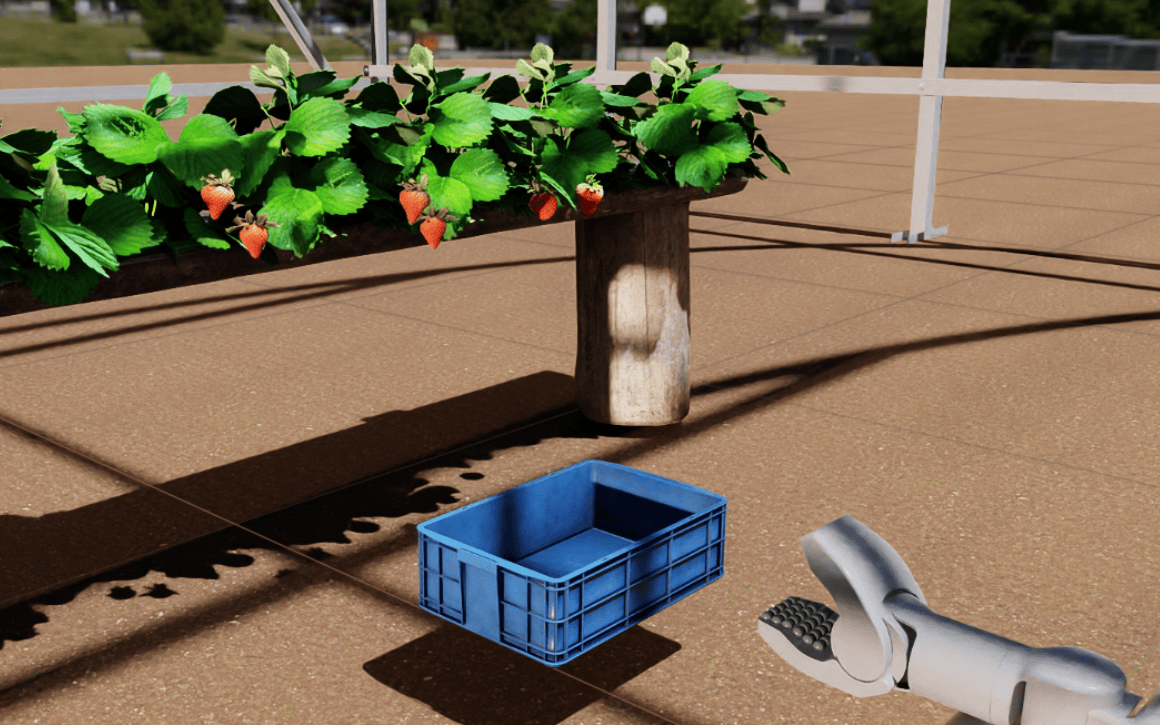} % Replace with actual simulation image file
    \end{minipage}
    \caption{Simulation and real-world environments for data collection. (Left) Real-world lab setup replicating an indoor greenhouse strawberry plantation. (Right) Simulated strawberry field in NVIDIA Isaac Sim}
    \label{fig:sim_real_envs}
\end{figure*}

For real-world data collection, we captured RGB-D data using an Intel RealSense D435i camera in our lab environment, designed to replicate an indoor greenhouse strawberry plantation. The lab setup includes five hanging strawberries with varied arrangements to simulate realistic harvesting conditions. The collected RGB-D data were processed to generate partial point clouds, which were manually annotated for ground truth labeling and combined with the simulation dataset to enhance model robustness. Both the simulation and real-world environments, including the camera perspectives and strawberry field layouts, are shown in Fig. \ref{fig:sim_real_envs}.

\subsection{Robotic System}
\begin{figure}[tb]
    \centering
    \includegraphics[width=0.8\columnwidth]{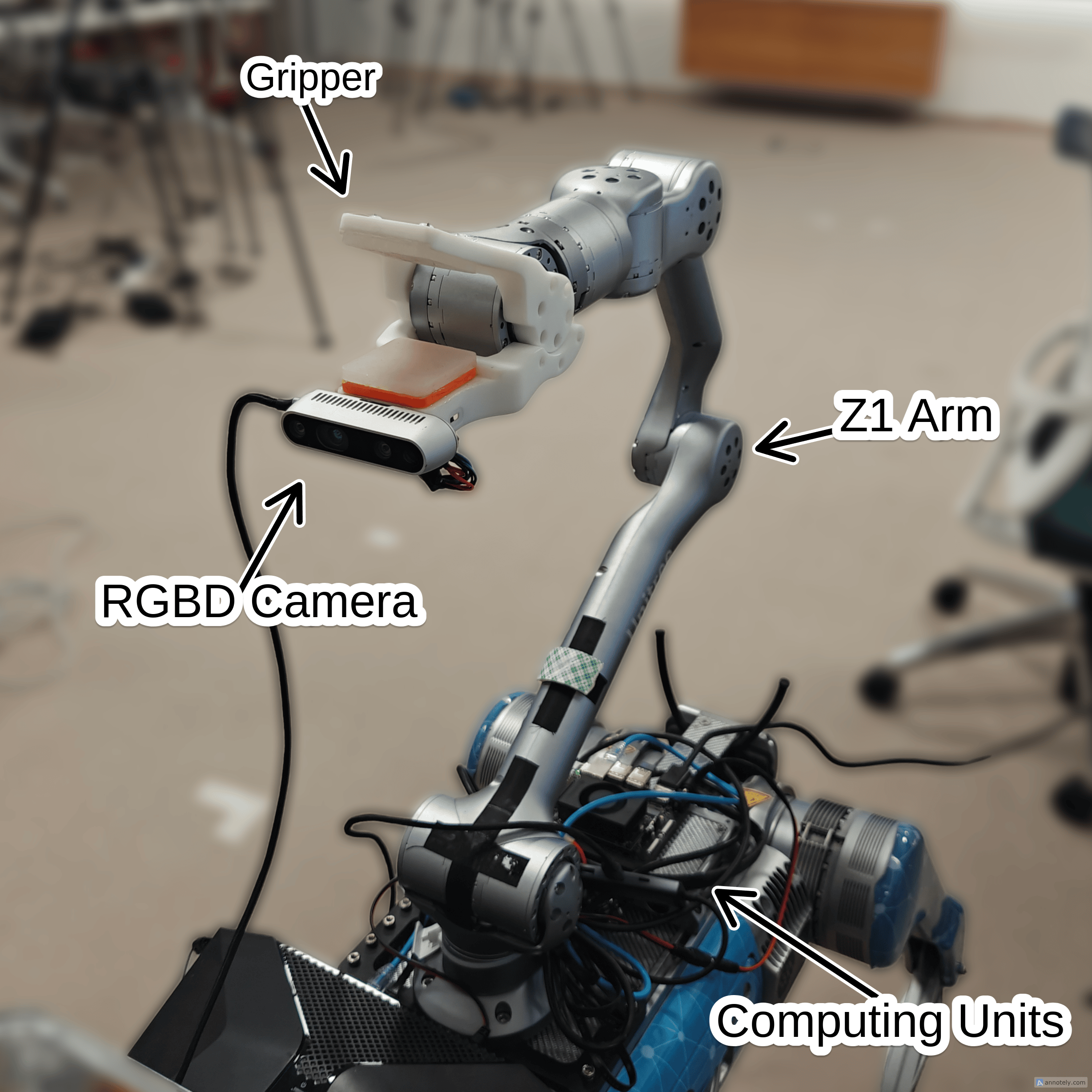} % Replace with actual image file
    \caption{Robotic system for strawberry harvesting. The image shows the Unitree Z1 robotic arm, grasping mechanism, Intel RealSense D435i RGB-D camera, and NVIDIA Jetson AGX Orin, with annotated components.}
    \label{fig:robotic_system}
\end{figure}

Once validated in simulation, we deployed our approach on a real-world robotic system designed to replicate an indoor greenhouse strawberry plantation. Our robotic platform consists of a Unitree Z1 robotic arm equipped with a gripper and an Intel RealSense D435i RGB-D camera. The camera is mounted on the robot to capture depth images of five hanging strawberries in a controlled environment. The robot's computational resources are powered by an NVIDIA Jetson AGX Orin. The robotic system is shown in Fig. \ref{fig:robotic_system}.

% Figure for robotic system with annotated components (single column)

% For ground truth labeling, we manually annotated 100 partial point clouds from five different recorded sequences, each containing varied arrangements of the strawberries in our setup. We recorded the transformation from the strawberry model to each partial point cloud and combined these with our simulation dataset to enhance robustness. After denoising the results, we used this dataset to train our point cloud completion model, PointAttn.

\subsection{Vision Modules}
To support our point cloud completion pipeline, we developed vision modules for strawberry detection and segmentation. First, we trained a  YOLOv8 \cite{yolov8_ultralytics} model for strawberry detection using RGB images collected from both the simulation and real-world environments. The training dataset was augmented with variations in lighting, occlusion, and strawberry orientations to improve generalization.

For segmentation, we utilized the Segment Anything Model 2 (SAM2) \cite{sam2} to generate precise masks for the detected strawberries. The masks were used to refine the partial point clouds, isolating strawberry-specific data for input into our point cloud completion model.

\subsection{Metrics}
To evaluate our system's performance, we conducted 80 real-world trials, using the completed point clouds for obstacle-aware motion planning. We employed several key metrics to assess performance: Attempts Ratio ($\rho_a$), which measures the ratio of grasp attempts to the number of detections; Success Rate ($\rho_s$), which calculates the ratio of successful grasps to detections; Success-to-Attempts Ratio ($\rho_s / \rho_a$), which indicates efficiency by showing how many attempts lead to successful grasps; Chamfer Distance (CD), used as a loss function to measure geometric accuracy of reconstructed point clouds against ground truth data; and Hit Rate ($\rho_h$), representing the ratio of incidents where other fruits in the scene were hit to the total number of grasp attempts.

\subsection{Baselines}

We compare our approach's success rate to the work in \cite{3d_shape_completion}, which shares some similarities with our overall approach. Both use point cloud processing, but our objectives differ: our approach specifically aims to grasp strawberries without damaging surrounding fruits in cluttered environments, whereas theirs focuses on individual strawberry grasping. Their setup is based on a vertical farming system and ours on a horizontal one, though the mathematical formulation of the point cloud completion remains similar. Both approaches take a partial point cloud as input and generate a completed point cloud as output, but our method uniquely accounts for neighboring strawberries to enable precise grasping in dense settings.

For chamfer distance evaluation, we compare against \cite{pan2023panoptic}, which utilizes a pre-trained DeepSDF model to encode the strawberry's shape representation and employs an occlusion-aware differentiable rendering pipeline for shape completion and pose estimation, as well as \cite{magistri2022contrastive}, which also focuses on reconstructing strawberries in a controlled environment using alternative methods. These approaches provide valuable benchmarks for assessing the accuracy and robustness of our reconstruction framework.

% While their method reconstructs strawberries to enhance fruit recognition and segmentation, ours models them as obstacles to enable safe grasping while ensuring minimal damage to surrounding fruits.

\section{Results and Discussions}
Our proposed pipeline effectively integrates point cloud completion with obstacle-aware trajectory planning, surpassing existing methods in both accuracy and robustness in strawberry harvesting.

\textbf{\textit{Grasping Success Rate:}} As shown in Table \ref{tab:sucess_rate_strawberry}, our method achieves the highest grasping success rate, with $\rho_s = 79.17\%$ and a precision ratio $\rho_s / \rho_a = 89.58\%$, outperforming all baselines. The Shape Completion method \cite{3d_shape_completion}, the closest competitor, achieves $\rho_s = 62.50\%$, highlighting a relative improvement of 26.67\%. This increase is attributed to our accurate pose estimation and obstacle-aware trajectory planning, which enables the robotic arm to execute more stable and precise grasps.

\begin{table}[tb]
\caption{Grasping Success Rate results on Strawberries.}
\label{tab:sucess_rate_strawberry}
    \centering
    \begin{tabular}{lccc}
        \toprule
        \textbf{Approach} & \multicolumn{3}{c}{\textbf{Strawberry}} \\
        & $\rho_a \uparrow$ [\%] & $\rho_s \uparrow$ [\%] & $\rho_s / \rho_a \uparrow$ [\%] \\
        \midrule
        % Grasp Point Detection \cite{grasp_point_detection} & 31.25 & 21.88 & 70.00 \\
        % Rotation Estimation \cite{rotation_estimation} & 78.13 & 40.63 & 52.00 \\
        % Ellipsoid Fitting \cite{ellipsoid_fitting} & 68.75 & 37.50 & 54.55 \\
        Shape Completion \cite{3d_shape_completion} & 84.38 & 62.50 & 74.07 \\
        \textbf{Ours} & \textbf{88.37} & \textbf{79.17} & \textbf{89.58} \\
        \bottomrule
    \end{tabular}
\end{table}

\textbf{\textit{Shape reconstruction accuracy:}} Table \ref{tab:chamfer_distance_strawberry} presents the Chamfer Distance (CD) results, where our method, using PointAttn for point cloud completion, achieves the lowest CD of 1.1 mm, substantially outperforming the best baseline (DeepSDF-based \cite{pan2023panoptic}) at 2.42 mm. This indicates that our approach generates more precise 3D shape reconstructions, which is crucial for reliable grasp planning. The improved shape completion ensures better localization of the strawberry's center and orientation, reducing grasping errors caused by incomplete or noisy point clouds.

\begin{table}[tb]
\caption{Chamfer Distance (CD) results on Strawberries}
\label{tab:chamfer_distance_strawberry}
\centering
\begin{tabular}{l c}
\toprule
\textbf{Approach} & \textbf{$D_C [mm] \downarrow$ avg} \\
\midrule
% DeepSDF \cite{park2019deepsdf} & 3.61 \\
CoRe \cite{magistri2022contrastive} & 2.67 \\
DeepSDF-based \cite{pan2023panoptic} & 2.42 \\
\textbf{Ours} & \textbf{1.10} \\
\bottomrule
\end{tabular}
\end{table}

\textbf{\textit{Obstacle Avoidance and Grasp Quality:}} A key advantage of our method is its ability to avoid non-target strawberries, as demonstrated in Table \ref{tab:success_and_hit_rate_strawberry}. When our method is enabled, the obstacle hit rate $\rho_h$ drops dramatically from 43.33\% to 13.95\%, representing a 67.8\% reduction. This indicates that our approach significantly improves safety and efficiency in real-world scenarios by minimizing damage to surrounding fruits while maximizing successful harvests.

\begin{table}[t]
\caption{Grasping Success Rate and Obstacle Hit Rate results on Strawberries.}
\label{tab:success_and_hit_rate_strawberry}
    \centering
    \begin{tabular}{ccc}
        \toprule
        \textbf{Ours Used} & $\rho_s / \rho_a \uparrow$ & Obstacle Hit Rate $\rho_h \downarrow$ [\%] \\
        \midrule
        $\times$ & 72.32 & 43.33 \\
        $\checkmark$ & \textbf{89.58} & \textbf{13.95} \\
        \bottomrule
    \end{tabular}
\end{table}

\section{Conclusions and future work}

In this paper, we presented a novel robotic strawberry harvesting pipeline that effectively addresses the complex challenges of operating in cluttered agricultural environments by integrating robust point cloud processing with complete 3D reconstruction and obstacle-aware trajectory planning to achieve significant performance improvements in real-world scenarios. The core of our method lies in three technical innovations: our denoising and segmentation techniques that reduce sensor noise and bridge the sim-to-real gap, our point cloud completion model that reconstructs strawberries from single-view RGB-D inputs, and finally a planning framework which treats non-target strawberries as obstacles. Experimental results validate the effectiveness of our approach, demonstrating substantial improvements over existing methods across all performance metrics, with markedly higher grasping success rates while maintaining excellent precision, superior shape reconstruction accuracy with significantly lower Chamfer Distance enabling more reliable grasp planning, and dramatically reduced contact with non-target strawberries, greatly enhancing harvesting safety and efficiency. Despite these promising results, several challenges remain for future work, including incorporating temporal information through multiple viewpoints or active vision strategies, exploring adaptive shape completion methods that can handle greater variability in strawberry sizes, and investigating the integration of this pipeline with different end-effector designs and applications to other high-value crops with similar harvesting challenges. The techniques developed in this work have broader implications beyond strawberry harvesting, potentially benefiting various agricultural robotics applications where precise manipulation in cluttered environments is required, representing a significant step toward more efficient, damage-free automated harvesting systems for delicate crops.
% A conclusion section is not required. Although a conclusion may review the main points of the paper, do not replicate the abstract as the conclusion. A conclusion might elaborate on the importance of the work or suggest applications and extensions. 

%\addtolength{\textheight}{-12cm}   % This command serves to balance the column lengths

% \section*{APPENDIX}

% Appendixes should appear before the acknowledgment.

% I confirmed existing IROS papers don’t have references on a new page 
%\newpage
\bibliographystyle{IEEEtran}
% \bibliography{main.bib}
% Generated by IEEEtran.bst, version: 1.14 (2015/08/26)

\end{document}